%% file: main.tex
\documentclass[11pt,letterpaper]{article}

\usepackage{authblk}
\usepackage{naaclhlt2018/naaclhlt2018}

\usepackage{color}
\usepackage{times}
\usepackage{url}
\usepackage{latexsym}
\usepackage{amsmath}
\usepackage{graphicx}
\usepackage{tabularx}
\usepackage{booktabs}
\usepackage[framemethod=tikz]{mdframed}
\usepackage{subcaption}

\usepackage{hyperref}

\usepackage{fancyvrb}
\usepackage{mathrsfs}
\usepackage{multirow,makecell}
\usepackage{pmboxdraw}
\usepackage{enumitem}
\usepackage{natbib}

\usepackage{enumitem}

\usepackage{tikz}
\usepackage{tikz-dependency}
\usepackage{pgfplots}
\usepackage{pgfplotstable}
\usepgfplotslibrary{groupplots}

\usepackage{abbrevs}
\newabbrev\QA{question-answer (QA)}[QA]
\newabbrev\PAS{predicate-argument structure (PAS)}[PAS]
\newabbrev\PA{predicate-argument}[predicate-argument]

\pgfplotsset{
  tick label style = {font=\scriptsize},
  every axis label = {font=\footnotesize},
  legend style = {font=\footnotesize},
  label style = {font=\footnotesize},
  compat=1.13
}
\usetikzlibrary{positioning}
\usetikzlibrary{patterns}
\usetikzlibrary{trees}
\usetikzlibrary{plotmarks}

\aclfinalcopy

\def\numWikiTrainSentences{3,938}
\def\numWikiTrainQAs{73,561}

\def\numGoodWikiTrainQAs{51,063}

\def\numWikiDevSentences{499}
\def\numWikiDevQAs{27,535}

\def\numGoodWikiDevQAs{19,069}

\def\numWikiTestSentences{480}
\def\numWikiTestQAs{26,994}

\def\numGoodWikiTestQAs{18,959}

\def\numPTBSentences{253}
\def\numPTBQAs{27,082}

\def\numGoodPTBQAs{18,789}

\def\numTotalQAs{155,122}
\def\numValidTotalQAs{132,660}
\def\pctValidTotalQAs{85.52\%}
\def\numGoodTotalQAs{107,880}
\def\pctGoodOfValidTotalQAs{81.3\%}

\def\ptbCost{\$2,862}
\def\wikiTrainCost{\$7,879}
\def\wikiDevCost{\$2,919}
\def\wikiTestCost{\$2,919}

\def\ptbCostPerToken{\$0.44}
\def\wikiTrainCostPerToken{\$0.08}
\def\wikiDevCostPerToken{\$0.25}
\def\wikiTestCostPerToken{\$0.25}

\title{Crowdsourcing Question-Answer Meaning Representations}

\author[1]{Julian Michael}
\author[2]{Gabriel Stanovsky}
\author[1]{Luheng He}
\author[2]{Ido Dagan}
\author[1,3]{Luke Zettlemoyer}
\affil[1]{Paul G. Allen School of Computer Science $\&$ Engineering, University of Washington, Seattle, WA}
\affil[2]{Bar-Ilan University Computer Science Department, Ramat Gan, Israel}
\affil[3]{Allen Institute for Artificial Intelligence, Seattle, WA}
     
\affil[  ]{\tt \{julianjm,luheng,lsz\}@cs.washington.edu}
\affil[ ]{{\tt gabriel.stanovsky@gmail.com}, {\tt dagan@cs.biu.ac.il}}
\date{}

\begin{document}
\maketitle

\begin{abstract}
We introduce Question-Answer Meaning Representations (QAMRs), which represent the predicate-argument structure of a sentence as a set of question-answer pairs.
We also develop a crowdsourcing scheme to show that QAMRs can be labeled with very little training, and gather a dataset with over 5,000 sentences and 100,000 questions.
A detailed qualitative analysis demonstrates that the crowd-generated question-answer pairs cover the vast majority of predicate-argument relationships in existing datasets (including PropBank, NomBank, QA-SRL, and AMR) along with many previously under-resourced ones, including implicit arguments and relations. 
The QAMR data and annotation code is made publicly available\footnote{\url{https://github.com/uwnlp/qamr}} to enable future work on how best to model these complex phenomena.

\end{abstract}

\input{introduction}
\input{collection}

\input{analysis}
\input{structure}
\input{related}
\input{future}


\bibliography{references}
\bibliographystyle{naaclhlt2018/acl_natbib}

\end{document}

%% file: introduction.tex
\section{Introduction}
\label{sec:intro}

Predicate-argument relationships form a key part of sentential meaning representations, and support answering basic questions such as \textit{who did what to whom}. 
Resources for predicate-argument structure are well-developed for verbs (e.g. PropBank~\cite{palmer2005proposition} and FrameNet~\cite{baker1998berkeley}) and there have been efforts to study other parts of speech (e.g. NomBank~\cite{meyers2004nombank}) and introduce whole-sentence structures (e.g.  AMR~\cite{banarescu2012abstract}).
However, highly skilled and trained annotators are required to label data within these formulations for each new domain, and it takes significant effort to model each new type of relationship (e.g., noun arguments in NomBank). We propose a new method to annotate relatively complete representations of the predicate-argument structure of a sentence, which can be done easily by non-experts.

\input{figures/intro_figure}

We introduce Question-Answer Meaning Representations (QAMRs), which represent the predicate-argument structure of a sentence with a set of question-answer pairs. Figure~\ref{fig:intro}
shows an example QAMR. Following the QA-SRL formalism~\cite{he2015qasrl}, each question-answer pair corresponds to a predicate-argument relationship.
There is no need for a carefully curated ontology of possible relationships and the labels are highly interpretable.
However, we differ from QA-SRL in focusing on \textit{all words} in the sentence (not just verbs) and allowing \textit{free form} questions (instead of using question templates).


One key advantage of this approach is that QAMRs can be annotated with crowdsourcing.
The biggest challenge is coverage, as it can be difficult to get a single annotator to write all possible QA pairs for a sentence.
Instead, we introduce a novel crowdsourcing scheme, detailed in \autoref{sec:collection}, where we distribute the work over multiple annotators by focusing them on different parts of the sentence, and include a simple incentive mechanism to encourage the labeling of as many QA pairs as possible.
In total, we gather over 100,000 questions and their answers for 5,000 sentences in two domains (Newswire and Wikipedia).

Although the free-form nature of QAMR questions is crucial for efficient labeling of meaning relationships for all words, it also presents a unique challenge:
The exact correspondence between QA pairs and predicate-argument relations is not obvious.\footnote{QA-SRL, while it does label relations with natural language questions, solves this problem by restricting the expressivity of the questions to verb-aligned templates.}
To study this challenge, in \autoref{sec:structure} we introduce
a simple algorithm to deterministically convert a QAMR into a graph that more closely resembles traditional formulations of predicate-argument structure.
These \textit{QAMR graphs} have overall structural similarities to AMR (with 80\% agreement on concept identification) and high coverage of relationships in PropBank and NomBank (including more than 90\% of non-discourse relationships).
However, because they are automatically derived from QAMRs, they can be easily gathered and are highly interpretable.

A qualitative analysis of the language in QAMRs, described in \autoref{sec:analysis}, reveals that they also capture a variety of phenomena that are not modeled in traditional representations of predicate-argument structure, including instances of coreference, implicit and inferred arguments, and implicit relations (for example, between nouns and their modifiers). Building effective models for such phenomena is an open challenge, and the QAMR data and annotation scheme should support significant future work in this direction.
Our code and data is open source and can be found at \url{https://github.com/uwnlp/qamr}.

%% file: figures/intro_figure.tex
\newmdenv[innerlinewidth=0.5pt, roundcorner=2pt,linecolor=black,innerleftmargin=6pt,
innerrightmargin=6pt,innertopmargin=6pt,innerbottommargin=6pt]{examplebox}

\begin{figure}[t!]
\small
\centering
\begin{examplebox}
Pierre Vinken, 61 years old, will join the board as a nonexecutive director Nov. 29. 
\vspace{5pt}
\textcolor{black}{\hrule height 0.6pt}
\vspace{5pt}
Who will \textbf{join} as \textbf{nonexecutive director}? - Pierre Vinken \\
What is \textbf{Pierre}'s last name? - Vinken \\
Who is \textbf{61 years old}? - Pierre Vinken \\
How \textbf{old} is \textbf{Pierre Vinken}? - 61 years old \\ 
What will he \textbf{join}? - the board \\
What will he \textbf{join the board} as? - nonexecutive director \\
What type of \textbf{director} will \textbf{Vinken} be? - nonexecutive \\
What day will \textbf{Vinken join the board}? - Nov. 29 
\end{examplebox}
\vspace{-1em}
\caption{Example QAMR annotation, where a set of question-answer pairs represents the predicate-argument structure of a sentence.}
\label{fig:intro}
\end{figure}

%% file: collection.tex
\section{Crowdsourcing}
\label{sec:collection}

\input{examples}



Our crowdsourcing task\footnote{Collected using the Amazon Mechanical Turk platform: \url{www.mturk.com}} is designed
with two principles in mind: gather questions that are as \textbf{simple} as possible, and be as \textbf{exhaustive} as possible over the sentence.
To achieve this, we used monetary incentives and crowd-arbitrated quality control in a two stage pipeline.
Workers in the first step (\textit{generation}) write questions and answers, and workers in the second step (\textit{validation}) answer or reject them.

\paragraph{Generation}


To ensure coverage of question generation, we show workers an English sentence with up to four underlined {\em target words}.
They are asked to write as many \textit{valid} QA pairs as possible containing each target word.
We define a \textit{valid question} to:
(1) contain at least one word from the sentence,
(2) be about the sentence's meaning, 
(3) be answered obviously and explicitly in the sentence,
(4) not be a yes/no question, and
(5) not be \textit{redundant}, where we define two questions as being redundant by the informal criterion of ``having the same meaning'' and the same answer.
The answers are (possibly non-contiguous) sets of tokens from the sentence.
All of these requirements were illustrated with examples in the instructions.

\paragraph{Validation}


The second component of our annotation pipeline is a question answering task. Workers receive a sentence (with no marked words) and a batch of questions that was written by an annotator in the first step (with no answers). For each question, the worker must mark it as \textit{invalid}, mark it as \textit{redundant} with another question, or highlight an answer in the original sentence, following the criteria above. Two workers validate and answer each set of questions. They are paid a base rate of 10c for each batch, with an extra 2c for each question they validated past four.


\paragraph{Incentives}
We use monetary incentives to increase coverage. In the generation task, a worker is required to write at least one \QA pair for each target word to receive the base pay of 20c. An increasing bonus of $3(k+1)$ c is paid for each $k$-th additional \QA pair they write that passes the validation stage.

\paragraph{Quality control}
For generation, workers are disqualified if the percentage of valid judgments on their questions falls below 75\%. 
For validation, workers need to pass a qualification test and maintain above 70\% agreement with others,
where answer spans are considered to agree if they have any overlap.


\subsection{Data Preparation and Annotation}
We drew our data from a set of 1,000 Wikinews articles from 2012--2015 and 1,000 articles from Wikipedia's list of 1,000 core topics.\footnote{\url{https://en.wikipedia.org/wiki/Wikipedia:1,000_core_topics}} 
We performed tokenization and sentence segmentation using the Stanford CoreNLP tools \cite{manning2014stanford} and partitioned each set by document into train, dev, and test.
We randomly sampled paragraphs from each part to gather about 2,000 sentences from each domain in the training set, and 250 from each in the dev and test sets.
We also annotated 253 sentences from the Penn Treebank training and dev sets, chosen to overlap with existing resources for comparison (see \autoref{sec:structure}).
For each sentence, we grouped its non-stopwords sequentially into groups of 3 or 4 to present to annotators as the target words, filtering out sentences that containing no content words.



\subsection{Dataset Statistics}
\begin{table}
\small
\newcolumntype{Y}{>{\centering\arraybackslash}X}
\begin{tabularx} {\columnwidth}{l c *{4}{Y}}
\toprule
    & \textbf{PTB}      & \textbf{Train}
    & \textbf{Dev}      & \textbf{Test}
    \\ 
    \midrule
Sentences 
    & \numPTBSentences{}
    & \numWikiTrainSentences{}
    & \numWikiDevSentences{}
    & \numWikiTestSentences{}
    \\ 
Annotators 
    & 5                         & 1
    & 3                         & 3
    \\ 
\QA Pairs
    & \numPTBQAs{} 
    & \numWikiTrainQAs{}
    & \numWikiDevQAs{} 
    & \numWikiTestQAs{}
    \\ 
Filtered 
   & \numGoodPTBQAs{}         & \numGoodWikiTrainQAs{}
   & \numGoodWikiDevQAs{}     & \numGoodWikiTestQAs{}
   \\ 
Cost
   & \ptbCost{}
   & \wikiTrainCost{}
   & \wikiDevCost{}
   & \wikiTestCost{}
   \\
Cost/token
   & \ptbCostPerToken{}
   & \wikiTrainCostPerToken{}
   & \wikiDevCostPerToken{}
   & \wikiTestCostPerToken{}
   \\ 
   \bottomrule
    \\ 
\end{tabularx}
\caption{Summary of the amount and cost of data gathered.}
\label{tab:data-summary}
\vspace{-1em}
\end{table}

\begin{figure}
    \newcolumntype{Y}{>{\centering\arraybackslash}X}
    \includegraphics{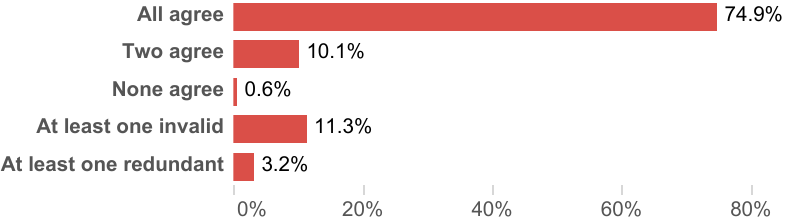}
    \caption{Agreement and validation statistics on all data gathered. Answers were considered to agree if their spans overlapped.}
    \label{fig:agreement}
\end{figure}

We gathered 1 annotation for each batch in the training set, 3 for each batch in the dev and test sets, and 5 for each batch on the Penn Treebank sentences, constructing a relatively large training set and smaller but more exhaustive development/test/comparison sets.
In total, we collected \numTotalQAs{} questions with three answer judgments each. \numValidTotalQAs{} (\pctValidTotalQAs{}) of these questions were rated as valid by both validators.  \autoref{fig:agreement} shows overall agreement statistics for question validation, which were very high across the three judgements.
After filtering out questions counted invalid or redundant by either annotator, we also remove questions not beginning with a wh-word,\footnote{\textit{who, what, when, where, why, how, which,} and \textit{whose}} which we found to be low-quality. This left \pctGoodOfValidTotalQAs{} of the valid questions, yielding \numGoodTotalQAs{} filtered questions in total.
See \autoref{tab:data-summary} for the breakdown of these numbers across the dataset.

\def\numUniqueQuestions{63,790}
\def\numQuestionsAppearingOnce{59,697}
\def\pctUniqueQuestionsAppearingOnce{94\%}
\def\pctCoveredByQuestionsAppearingOnce{85\%}

%% file: examples.tex
\begin{table*}[t]
\scriptsize
\newcolumntype{Y}{>{\arraybackslash}X}
\newcolumntype{S}{>{}X}
\setlength{\tabcolsep}{.4em}
\begin{tabularx}{\textwidth}{S l* {2}{Y} X}
\toprule
\textbf{Sentence} & \textbf{Ann.}  & \textbf{Question}  & \textbf{Answers} \\\midrule
\multirowcell{2}[0ex][l]{
(1) Climate change affects distribution of\\ weeds, pests, and diseases.}
            &
            & (a) What affects distribution of diseases?  
            & Climate change \\
            & VAR	
            & (b) What is affected? 
            & distribution of... / 	distribution \\
\cmidrule(lr){1-4}
\multirowcell{3}[0ex][l]{
(2) Baruch ben Neriah, Jeremiah's scribe,\\ used this alphabet to create the later \\ scripts of the Old Testament.}
            & SYN
            & (a) Who wrote the scripts?
            & Baruch ben Neriah \\
            & ROLE
            & (b) Who did Baruch work for?
            & Jeremiah \\
            &
            & (c) What is old?
            & Testament / the Old Testament \\
\cmidrule(lr){1-4}
\multirowcell{3}[0ex][l]{
(3) Mahlunga has said he did nothing \\ wrong and Judge Horn said he ``failed to \\ express genuine remorse''.}
            & ROLE
            & (a) What is the Judge's last name?	
            & Horn \\
            & INF
            & (b) Who doubted his remorse was genuine?
            & Judge Horn \\
            & CO
            & (c) Who didn't express genuine remorse?
            & Mahlunga \\
\cmidrule(lr){1-4}
\multirowcell{3}[0ex][l]{
(4) Cyclones of this level of intensity (as\\ measured by top wind speed and central\\ pressure value) are very, very rare.}
            &
            & (a) What with this level of intensity are rare?	
            & Cyclones \\
            &
            & (b) What kind of speed is it?	
            & wind / top \\
            &
            & (c) What about it is very rare?
            & level	/ this level of intensity \\
\cmidrule(lr){1-4}
\multirowcell{3}[0ex][l]{
(5) Smaller snakes eat smaller prey.}
            & 
            & (a) What eats?	
            & snakes \\
            & 
            & (b) What do smaller snakes do?
            & eat / eat smaller prey \\
            & 
            & (c) What size snakes?
            & Smaller \\
\cmidrule(lr){1-4}
\multirowcell{3}[0ex][l]{
(6) In Byron's later memoirs, ``Mary \\ Chaworth is portrayed as the first object \\ of his adult sexual feelings.''}
            & 
            & (a) Who is portrayed in the work?
            & Mary Chaworth \\
            & IMP
            & (b) Who was the object of his sexual feelings?
            & Mary Chaworth \\
            & VAR
            & (c) Who was Mary the object of sexual feelings for?
            & Byron \\
\cmidrule(lr){1-4}
\multirowcell{4}[0ex][l]{
(7) Volunteers are presently renovating \\ the former post office in the town \\ of Edwards, Mississippi, United States \\ for the doctor to have an office.}
            & 
            & (a) What town is the post office in?
            & Edwards \\
            & 
            & (b) What state is the post office in?
            & Mississippi \\
            & IMP
            & (c) What country are the volunteers renovating in?
            & United States \\
            & VAR
            & (d) What country is the city of Edwards in?
            & United States \\
\cmidrule(lr){1-4}
\multirowcell{2}[0ex][l]{
(8) The ossicles are the malleus (hammer), \\ incus (anvil), and the stapes (stirrup).}
            & VAR
            & (a) What is the malleus one of?
            & The ossicles / ossicles \\
            & & & \\
\cmidrule(lr){1-4}
\multirowcell{4}[0ex][l]{
(9) Liam ``had his whole life in front \\ of him'', said Detective Inspector \\ Andy Logan, who was the senior \\ investigator of his murder.}
            & ROLE
            & (a) Who is the murder victim Logan is investigating?
            & Liam \\
            & ROLE
            & (b) What rank of investigator is Andy Logan?
            & Detective Inspector / senior \\
            & INF
            & (c) Who was Detective Logan speaking about?
            & Liam \\
\cmidrule(lr){1-4}
\multirowcell{4}[0ex][l]{
(10) This cemetery dates from the time of \\ Menkaure (Junker) or earlier (Reisner), \\ and contains several stone-built mastabas \\ dating from as late as the 6th dynasty.}
            & INF
            & (a) How old are the stone-built mastabas?
            & dating from as late as the 6th dynasty / from as late as the 6th dynasty \\
            & IMP
            & (b) What period was earlier than Menkaure?
            & Reisner \\
            &
            & (c) What dates from the 6th dynasty?
            & mastabas / several stone-built mastabas \\

\bottomrule
\end{tabularx}
\caption{Examples of question-answer pairs capturing various semantic relations, annotated with interesting phenomena they exhibit: syntactic variation (VAR), synonym use (SYN), explicit role names for implicit relations (ROLE), coreference (CO), implicit arguments (IMP), and inferred relations (INF). Only distinct answers are listed.}
\label{tab:full-sample-examples}
\vspace{-1em}
\end{table*}

%% file: analysis.tex
\section{Data Analysis}
\label{sec:analysis}

\def\pctQuestionsBeginningWithWh{81.4\%}
\def\pctQuestionsNotBeginningWithWh{18.0\%}
\def\pctQuestionsNoWh{0.7\%}
\def\numQuestionsBeginningWithWh{70,132}

\def\pctWhat{60.9\%}
\def\pctWho{17.5\%}
\def\pctHow{6.9\%}
\def\pctWhere{5.0\%}
\def\pctWhen{4.3\%}
\def\pctWhich{2.9\%}
\def\pctWhose{1.9\%}
\def\pctWhy{0.6\%}

\begin{figure}[t]
    \centering
    \includegraphics[width=0.95\columnwidth]{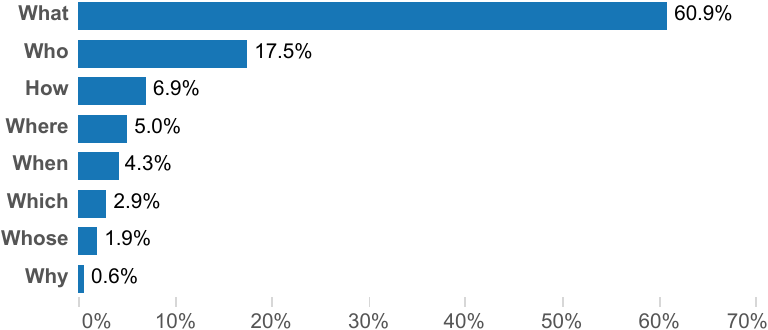}
    \caption{Distribution of wh-phrases among questions.}
\label{fig:wh-breakdown}
\vspace{-1em}
\end{figure}    

In this section, we perform a qualitative analysis on the questions and answers in the combined and filtered training and development sets, with a total of 70,132 \QA pairs.
We show that our open annotation format captures a wide variety of properties and relations, even using a rich vocabulary to label fine-grained semantic relations that are implicit in the original text.


\subsection{Variety of predicate-argument relations}

\def\numQAsRandomSampled{150}


We sampled a set of \QA pairs for analysis by randomly drawing sentences and 20\% of their annotated QA pairs until we reached \numQAsRandomSampled{} questions. 
See \autoref{tab:full-sample-examples} for example QA pairs which will serve as a reference for the remainder of this analysis.

We categorized our sample by the \PA relation they targeted in the original sentence.\footnote{We assume a QA pair targets the relation corresponding to the semantic role of the wh-word in the question.}
In over 90\% of cases, the \QA pair directly corresponded to a \PA relation expressed in the sentence.
These relations varied, including both traditionally annotated \PA structures for arguments and modifiers of nouns and verbs, as well as relationships within proper names (examples 2c, 3a) and coreference (examples 3c, 6c).

Finally, a small number of questions target relations which cannot be aligned to the sentence at all. In our sample, these were all shallow inferences over the sentence (examples 3b, 9c).

\subsection{Open-vocabulary types and roles}
\label{sec:open-vocab}
\def\externalWord{external phrase}
\def\ExternalWord{External phrase}

\begin{figure}[t]
\centering
    \includegraphics[width=0.9\columnwidth]{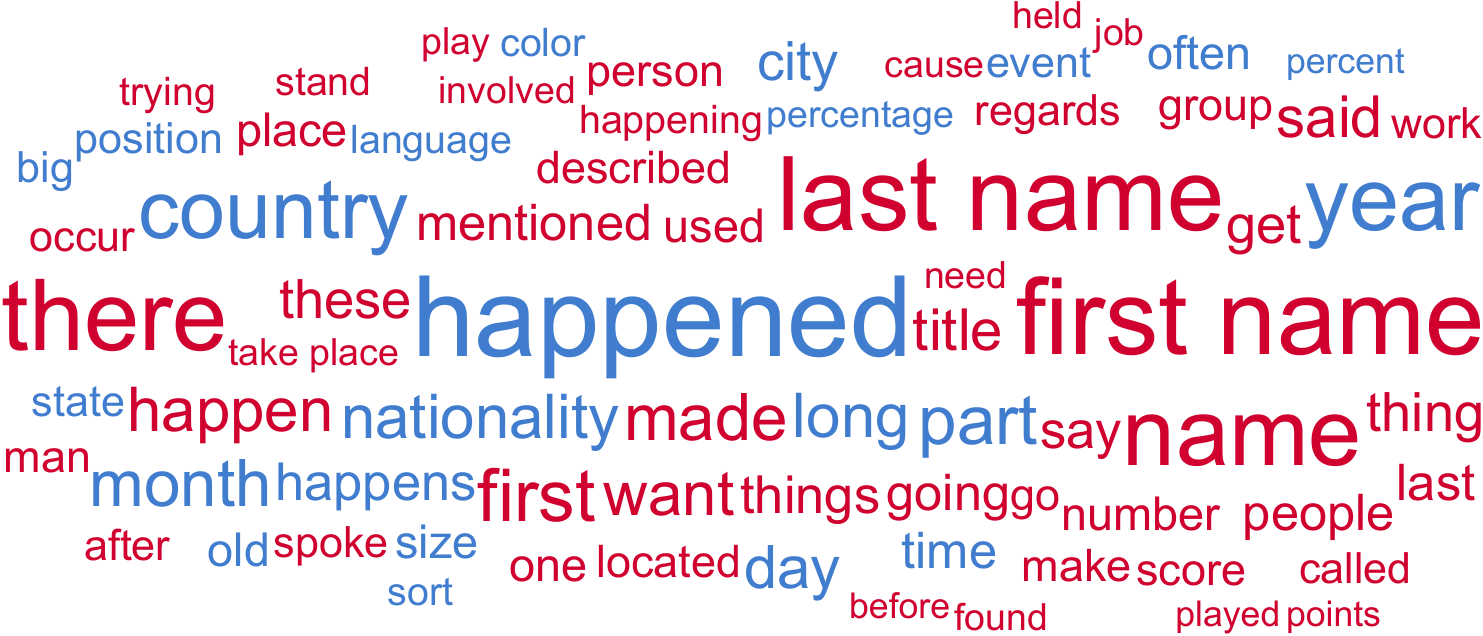}
\caption{External phrases appearing more than 50 times in questions. The most frequent one, \textit{happened}, occurred 746 times. Phrases that most commonly appear after \textit{who, which}, or \textit{how} are in (lighter) blue.}
    \label{fig:external-word-cloud}
    \vspace{-1em}
\end{figure}

Despite the strong correspondence with \PA structure from the original sentence, there was a large vocabulary of \textit{\externalWord{}}s---5,687 unique phrases appearing 25,952 times (excluding stopwords)---that appeared in questions but not their respective sentences.
In total, 38.7\% of questions contain an \externalWord{}.
See \autoref{fig:external-word-cloud} for the most common ones.

We sampled 120 QA pairs with \externalWord{}s in the question, and found that they provide a rich semantic characterization of the entities and relations involved in a sentence's \PA structure, as well as providing fine-grained identification of roles.


\paragraph{Entity typing}
In our sample, 42\% of all uses of \externalWord{}s were for typing entities.\footnote{Excluding cases of spelling errors and derivational morphology.}
Consider example 7 in Table \ref{tab:full-sample-examples}: the words \textit{state} and \textit{country} are both used in different questions to identify entities that were mentioned in the sentence.
This example also shows one reason these expressions were common: they were used to more clearly identify distinct answers to the questions where the semantic role being denoted (the location of the post office) was the same.
There is a broad class of typing words used: 1,105 unique \externalWord{}s appear directly after \textit{what}, \textit{which}, or \textit{how}, where they are used almost exclusively to type the answer.

\paragraph{Making implicit relations explicit}
Another common use of \externalWord{}s, making up 22\% of our sample, was to assign explicit names to semantic relations that were otherwise only expressed implicitly through syntax. The most common example: \textit{What is X's first name?} 
In Table \ref{tab:full-sample-examples}, example 2b, \textit{Baruch ben Neriah, Jeremiah's scribe} becomes \textit{Who did Baruch work for?}~and in example 9a, \textit{his murder} becomes \textit{Who was the murder victim Logan is investigating?}
There are also many examples that use \textit{how} to relate entities to gradable values: \textit{how long}, \textit{how often}, \textit{how old}, \textit{how big}, etc. (see example 10a).
Outside of our sample, we also found examples of verbs being used to paraphrase the relationships in noun compounds, similar to those proposed by \newcite{Nakov2008}. For example, the question \textit{Who conducted the poll?} for the phrase \textit{Gallup poll}, and \textit{Who received the bailouts?} for the phrase \textit{bank bailouts}.

There is room for improvement here: many noun modifiers are targeted with very general questions beginning with the phrase \textit{what kind} or \textit{what type} (e.g., example 4b).
In fact, those two prefixes alone count for 8.5\% of all questions.
However, in many cases it is easy to see how those relationships may be more richly characterized in the QAMR paradigm, for example, asking \textit{Which measurement of speed?~/ top} and \textit{What is it the speed of?~/ wind}.

\subsection{Semantics, not syntax}
Finally, these questions go far beyond reproducing the syntax of the original sentence.
By our analysis, only 63\% of \QA pairs characterize their predicate-argument relation using the same syntactic relationship as in the sentence.
5\% are answered with phrases coreferent with the actual syntactic argument (Table \ref{tab:full-sample-examples}, examples 3c, 6c);
17\% exhibit syntactic variation, using different prepositions (examples 6c, 8a), alternating between active and passive (example 1b), or changing between the noun and verb form of the predicate (example 10a);
6\% ask about implicit arguments that are not syntactically realized in the sentence (examples 6b, 7c, 10b);
and 6\% ask about inferred relations that are not explicitly stated in the sentence (examples 3b, 10b).


Finally, there are hints of richer semantic phenomena that could potentially be captured in the QAMR paradigm.
Example 4c, with the question \textit{What about it is very rare?} identifies the focus of the \textit{rare} predicate, while the questions in example 5 provide other clues about restrictivity and presupposition: while \textit{What eats?}~is only answered with \textit{snakes}, \textit{What do smaller snakes do?} may be answered with the more specific \textit{eat smaller prey}.
Finally, example 3c reveals the counter-factive nature of \textit{failed} in the phrase \textit{failed to express remorse}.

%% file: structure.tex
\section{Inducing QAMR Graphs}
\label{sec:structure}

\input{figures/structure_figure}



In this section, we present an algorithm to automatically convert QAMRs to simple, interpretable structures we call \textit{QAMR Graphs}.
In a quantitative comparison, we use this algorithm to show that QAMRs cover the vast majority of relevant relationships contained in other widely used resources, including PropBank, NomBank, and AMR, despite the fact that QAMR labels can be easily gathered from any native speaker.

\subsection{Definition}
Given a sentence $\mathbf{w} = w_1, \ldots, w_n$ and a QAMR $\{(q_i,a_i) \vert i=1\ldots m\}$, a \textit{QAMR Graph} is a graph whose nodes are subspans of $\mathbf{w}$ and whose edges are labeled with questions $q_i$. Each edge points from a span in $q_i$ to a span in the corresponding answer $a_i$, and none of the spans that appear as nodes intersect with each other. Figure~\ref{fig:struct} shows an example graph for the QAMR shown in Figure~\ref{fig:intro}.

\subsection{Algorithm}
\label{sec:struct_induction}

\paragraph{Step 1: Node identification}
    A span $S$ from the sentence is identified as a node in the QAMR graph if: (1)
    it appears contiguously within a question or an answer, 
    and (2) it is \emph{minimal}, in the sense that no sub-span of $S$ appears in any 
    other QA pair independently of $S$.
    For example, given the QA-pairs \\[5pt]
    (1) Who will [\textbf{join} \textbf{the board}]?  - [\textbf{Pierre}] \\ 
    (2) What will [\textbf{Pierre}] [\textbf{join}]? - [\textbf{the board}] \\[5pt]
    we extract the set \{join, Pierre, the board\}.

    \paragraph{Step 2: Predicate-argument extraction}
    After node identification,
    we assume that each QA pair denotes a single proposition involving exactly one predicate and a subset of its arguments, where the predicate appears in the question.
    To identify the question predicate,
    we first assign each node its {\em predicate score} $\frac{c_q}{c_q + c_a}$, where $c_q$ is the number of occurrences of the node in a question and $c_a$ is the number of occurrences in an answer.
    We then identify the highest-scoring node in each question as its predicate,
    and identify all other nodes in the QA pair (including those in the question) as its arguments.
    The question then acts as the edge label between the predicate and the answer for the purposes of step 3.

    \paragraph{Step 3: Structure Induction}
    Given a list of \PA structures from the previous step, we determine the final rooted graph structure by attaching all arguments to predicates by decreasing order of their predicate score and pruning non-projective edges from the result.
    
    In general, some of the resulting edges may come from a predicate attaching to arguments that appeared with it in a question, but not as an answer. These edges we give an empty label and treat as denoting relations \textit{missing} from the QAMR.
    

\subsection{Comparison to AMR}
To assess the quality of our results, we compare the results to Abstract Meaning Representation (AMR) structures.
We compute two of the metrics suggested by \newcite{damonte2016incremental}\footnote{We chose the metrics which are most relevant to our annotation, and do not require external knowledge, such as word sense disambiguation or wikification.}: (1)~Concept identification, which  measures agreement on the nodes of the structures, and (2)~an unlabeled variant of the SMATCH score \cite{cai2013smatch}, which measures the agreement on the set of relations between concepts, ignoring their label. We judge both metrics only on sentence elements, ignoring AMR relations such as entity typing and PropBank predicates with a different lemma than that which appeared in the surface form.

Results are shown in \autoref{tab:amr}. Overall, our method achieves good agreement with AMR, comparable with current 
state of the art AMR parsing (83\% F1 on concept agreement and 69\% F1 on unlabeled SMATCH).
These results confirm that our algorithm works well and affirm that QAMRs are effective at encoding the information present in more traditional representations of \PA structure.

\paragraph{Disagreements with AMR}

Analysis reveals that the lower precision for concept identification is due to
QAMR graphs' use of surface words to encode certain phenomena where AMR instead uses \textit{non-core roles}.
These are relations which AMR encodes as properties from a predefined closed semantic lexicon (roughly 50 roles), often requiring external knowledge to predict. 

Consider, for example, the word \textit{nonexecutive}.
AMR omits the morpheme \textit{non} and instead uses an \textit{executive} concept with a \emph{:polarity} modification, while the QAMR graph uses the original surface word as its concept.
In addition to negations (which account for 1.7\% of disagreements), this mismatch also occurs for phenomena such as numeric quantities (12.7\%),
date normalization (4.9\%), and pronoun resolution (2.3\%).
Further, 59\% of the missing relations according to the unlabeled SMATCH score stem from mismatch on non-core roles, such as \emph{:wiki}, \emph{:quant}, \emph{:consist-of}, and \emph{:part-of}.


\begin{table}[t]
\newcolumntype{Y}{>{\centering\arraybackslash}X}
\small
\centering
\begin{tabularx}{\columnwidth}{l * {4}{Y}}
\toprule
\textbf{Metric} & \textbf{Precision} & \textbf{Recall} & \textbf{F1} \\
\midrule
Concept agreement       & 70.0       & 94.0   & 80.2     \\
Unlabeled SMATCH         & 67.5     & 51.5        & 58.4 \\ 
\bottomrule
\end{tabularx}
\caption{Agreement of automatic QAMR graphs with gold AMR annotations on the first 100 sentences of the Penn Treebank development set.}
\label{tab:amr}
\vspace{-1em}
\end{table}

\subsection{Comparison to SRL}
\label{sec:comparisons}

In addition to the comparison to the structured representation of AMR, we  assess the coverage of QAMR on frame-based representations from three resources: PropBank, NomBank, and QA-SRL.


\paragraph{Preprocessing}
Each of these resources identifies a \PA structure as a set of labeled arcs from a head word to argument spans. For each, we consider only those \PA relationships within the scope of QAMR.

For PropBank, we filter out predicates and arguments that are auxiliary verbs, as well as reference ({\tt R-}) roles since aligning these properly is difficult and their function is primarily syntactic.
We also remove discourse (-DIS) arguments such as \textit{but} and \textit{instead}: these may be regarded as involved in \textit{discourse} structure separately from the \PA structure we are investigating. 78\% of the original dependencies remain.
For NomBank, we also remove auxiliaries, and we remove arguments that include the predicate---which are present for words like \textit{salesman} and \textit{teacher}---leaving 83\% of the original dependencies.
For QA-SRL, we use all dependencies, and where multiple answers were provided to a question, we take the union of the answer spans to be the argument span.

\paragraph{Methodology}
To compare the QAMR data to these resources, for each QA pair we apply the first two steps of the structure induction algorithm to determine its predicate, and and take its answer as the argument span.
Then, for each resource independently, we align each question answer pair to the predicate-argument arc with the highest relative overlap between the argument span and answer span such that the predicate is present in the question. This measures coverage of relations in these resources without relying on the details of span-finding decisions.

\paragraph{Results}

\begin{figure}
    \centering
    \includegraphics[width=0.9\columnwidth]{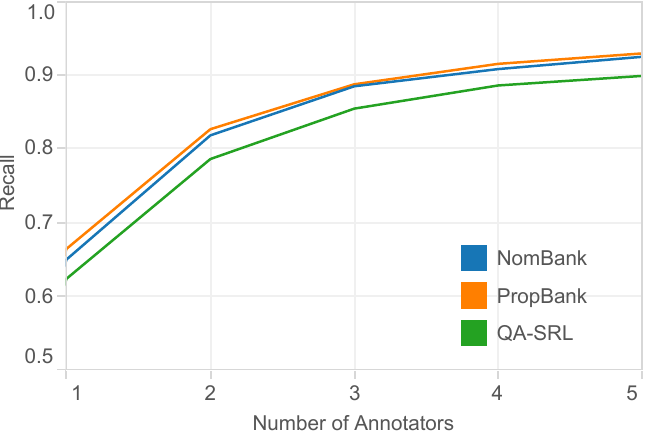}
    \caption{Recall of \PA relations for sentences shared with each of our reference datasets, with increasing number of annotators.}
    \label{fig:srl-comparison-recall-curve}
    \vspace{-1em}
\end{figure}

Of our 253 sentences from the Penn Treebank, 150 have QA-SRL labels, 223 have PropBank labels, and 232 have NomBank labels. 
We measure recall on these sentences in a series of simulated experiments by sampling $n$ annotators out of 5 for each grouping of target words, simulating the situation for the training set (1 annotator) and the dev/test sets (3 annotators).

The results are shown in \autoref{fig:srl-comparison-recall-curve}. Single annotators cover over 60\% of the relationships, and the number quickly increases as more annotations are provided by additional independent judgements. 

We manually examined 25 sentences for the 5-annotator case to study sources of coverage loss. 
In comparison to PropBank and NomBank, the missing dependencies are due to missing QA pairs (44\%), mistakes in our alignment heuristic (28\%), and subtle modifiers/idiomatic uses (28\%). For example, annotators sometimes overlook phrases such as \textit{so far} (marked as a temporal modifier in PropBank) or \textit{let's} (where \textit{'s} is marked as a core verbal argument). Comparing to QA-SRL, 60\% of the missed relations are inferred/ambiguous relations that are common in that dataset. Missed QA pairs in QA-SRL account for another 20\%. 

\subsection{Discussion}
In aggregate, these analyses show that the QAMR labels capture the same kinds of predicate-argument structures as existing resources. However, when only gathering annotations from one annotator for each target word (as in our training set), the recall on these relations is low compared to expert-annotated structures, which may pose challenges to learning. Improving the  annotation and structure induction (for example, by adding a second crowdsourcing stage to fill in the missing relations identified in step 3 of our structure induction algorithm) is an interesting avenue of future work.

%% file: figures/structure_figure.tex
\begin{figure*}[tb!]
\small
        \centering
  \resizebox{1\textwidth}{!}{
      \begin{dependency}[theme=simple]
        \begin{deptext}[column sep=0.2cm]
          Pierre \& Vinken \&, \& 61 \& years \& old \& , \& will \& join \& the \& board \& as \& a \& nonexecutive \& director \& Nov. \& 29 \\
        \end{deptext}
        \deproot{9}{ROOT}
        \depedge[edge end x offset=-3pt]{9}{1}{Who will join as nonexecutive director?}
        \depedge[label style={shift={(+0.58em,0em)}}]{1}{2}{What is Pierre's last name?}
        \depedge[edge end x offset=1pt]{6}{1}{Who is 61 years old?}
        \depedge[label style={shift={(-0.7em,0.07em)}}]{6}{4}{How old is Pierre Vinken?}
        \wordgroup{1}{4}{5}{years}
        \wordgroup{1}{16}{17}{nov}
        
        \wordgroup{1}{10}{11}{board}
        \depedge[edge end x offset=-17pt, label style={shift={(+0.4em,0em)}}]{9}{11}{What will he join?}
        \depedge[label style={shift={(+0.7em,0em)}}]{9}{15}{What will he join the board as?}
        \depedge[label style={shift={(-0.8em,0em)}}]{15}{14}{What type of director will Vinken be?}        
        \depedge{9}{16}{What day will Vinken join the board?}
  \end{dependency}}


    \caption{Automatically constructed QAMR graph for the first sentence of the PTB.}
    \label{fig:struct}
    \vspace{-.3em}
  \end{figure*}

%% file: related.tex
\section{Related Work}
\label{sec:related}

In addition to the semantic formalisms \cite{palmer2005proposition,meyers2004nombank,banarescu2012abstract,he2015qasrl} we have already discussed, FrameNet \cite{baker1998berkeley} also focuses predicate-argument structure, but has more fine-grained argument types. \newcite{gerber2010beyond} target implicit nominal arguments. \newcite{2016stanovskyRestrictive} annotate non-restrictive noun phrase modifiers on top of QA-SRL. Other linguistically motivated annotation schemes include UCCA \cite{abend2013universal},
HSPG treebanks~\cite{flickinger2017sustainable},
and the Groningen meaning bank \cite{basile2012developing}. 

Crowdsourcing has also been applied to gather annotations of structure in the setup of multiple choice questions, for example,
to recover Dowty's semantic proto-roles \cite{reisinger2015semantic,white2016universal} and for human-in-the-loop parsing and classification~\cite{he2016human,duan2016generating,werling2015job}.
\newcite{Wang2017crowd} use crowdsourcing with question-answer pairs to annotate some PropBank roles directly.
The paraphrases of implicit relations that arose in our annotations are closely related to the approach of \newcite{Nakov2008}, which used crowdsourcing to paraphrase the semantic relations in noun compounds.

Question-answering tasks such as SQuAD \cite{rajpurkar2016squad}, MCTest \cite{richardson2013mctest}, and VQA \cite{antol:2015:vqa} also use crowdsourcing for cheap and scalable collection of question-answer pairs, albeit for very different end purposes.





%% file: future.tex
\section{Conclusion and Future Work}
\label{sec:future}
QAMR provides a new way of thinking about meaning representations: using open-ended, natural language annotation to represent rich semantic structure.
This paradigm allows for representing a broad range of semantic phenomena with data that can be easily gathered from any native speaker.
How to best model these phenomena is an open challenge, which our annotation scheme could support studying at a relatively large scale.
We have prioritized predicate-argument structure, but in future work the approach may be extended to other semantic phenomena, multi-sentence contexts, and larger scales with human-in-the-loop or on-the-job learning.
